\documentclass[10pt,twocolumn,letterpaper]{article}

\usepackage{cvpr}              %

\newcommand{\refsec}[1]{Sec.~\ref{sec:#1}}

\newcommand{\reftbl}[1]{Tab.~\ref{tab:#1}}
\newcommand{\reffig}[1]{Fig.~\ref{fig:#1}}

\newcommand{\lblfig}[1]{\label{fig:#1}}
\newcommand{\lblsec}[1]{\label{sec:#1}}

\newcommand{\lbltbl}[1]{\label{tab:#1}}
\newcommand{\lblalg}[1]{\label{alg:#1}}

\newif\ifarxivversion
\arxivversiontrue %

\usepackage{bm}
\usepackage{amssymb}%
\usepackage{pifont}
\newcommand{\cmark}{\ding{51}}%
\newcommand{\xmark}{\ding{55}}%

\usepackage[dvipsnames]{xcolor}

\definecolor{greenkmeans}{RGB}{56,118,29}
\definecolor{bluekmeans}{RGB}{164,194,244}

\definecolor{cvprblue}{rgb}{0.21,0.49,0.74}
\usepackage[pagebackref,breaklinks,colorlinks,citecolor=cvprblue]{hyperref}
\usepackage[ruled,lined,linesnumbered]{algorithm2e}

\def\paperID{****} %
\def\confName{CVPR}
\def\confYear{2024}

\title{Streaming Dense Video Captioning}

\author{Xingyi Zhou\thanks{Equal contribution. \{zhouxy, aarnab\}@google.com} \quad Anurag Arnab\footnotemark[1] \quad Shyamal Buch \quad Shen Yan \\ Austin Myers \quad Xuehan Xiong \quad Arsha Nagrani  \quad  Cordelia Schmid 
\\
Google
}

\begin{document}
\maketitle
\begin{abstract}

An ideal model for dense video captioning -- predicting captions localized temporally in a video -- should be able to handle long input videos, predict rich, detailed textual descriptions, and be able to produce outputs before processing the entire video. Current state-of-the-art models, however, process a fixed number of downsampled frames, and make a single full prediction after seeing the whole video. We propose a streaming dense video captioning model that consists of two novel components: First, we propose a new memory module, based on clustering incoming tokens, which can handle arbitrarily long videos as the memory is of a fixed size. Second, we develop a streaming decoding algorithm that enables our model to make predictions before the entire video has been processed. Our model achieves this streaming ability, and significantly improves the state-of-the-art on three dense video captioning benchmarks: ActivityNet, YouCook2 and ViTT. Our code is released at \href{https://github.com/google-research/scenic/tree/main/scenic/projects/streaming_dvc}{https://github.com/google-research/scenic}.
\end{abstract}

\section{Introduction}

\begin{figure}[t]
    \center
    \vspace{-\baselineskip}
    \begin{subfigure}{\linewidth}
        \includegraphics[width=0.99\linewidth, page=1]{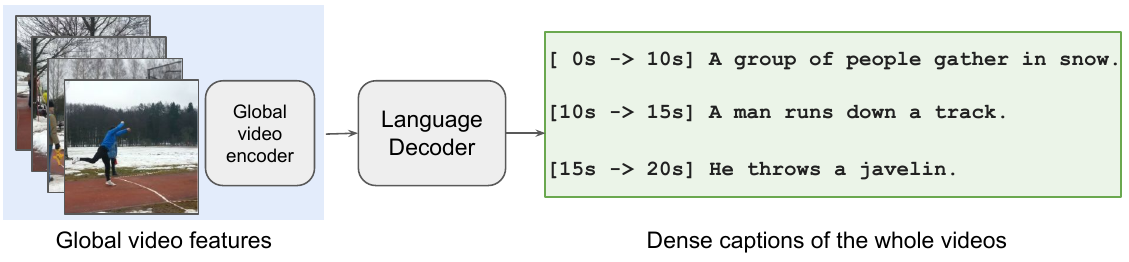}
        \caption{
            \textbf{Conventional global dense video captioning models}~\cite{yang2023vid2seq,wang2021end}.}
    \end{subfigure}
    \begin{subfigure}{\linewidth}
        \includegraphics[width=0.99\linewidth, page=2]{figures/Streaming_DVC_teaser_v4.pdf}
        \caption{\textbf{Our streaming dense video captioning model.}
    }
    \vspace{-3mm}
    \end{subfigure}
	\caption{\textbf{Comparing our streaming model (b) to conventional global models (a).}
    Conventional global models encode the entire video at once, and produce captions for all events at the end.
    Our model encodes images frame-by-frame, uses them to update a running memory, and predicts captions sequentially. 
    }
    \lblfig{teaser}
    \vspace{-\baselineskip}
\end{figure}

Video is ubiquitous in modern society, quickly becoming one of the most prevalent media formats for transmitting information.
The majority of computer vision models designed for video understanding only process a handful of frames, mostly covering only a few seconds~\cite{ wang2023videomae, ryali2023hiera, yan2022multiview, lin2022frozen, li2022uniformerv2}, and are typically limited to classifying these short segments into a fixed number of concepts.
In order to achieve a comprehensive, fine-grained video understanding, we study the task of dense video captioning
-- jointly localizing events temporally in video and generating captions for them.
Ideal models for this goal should be able to handle both long input sequences -- to reason over long, untrimmed videos -- and also to handle long output sequences in text space, to describe in detail all the events within the video.

\begin{figure*}[t]
	\center
    \includegraphics[width=0.9\linewidth]{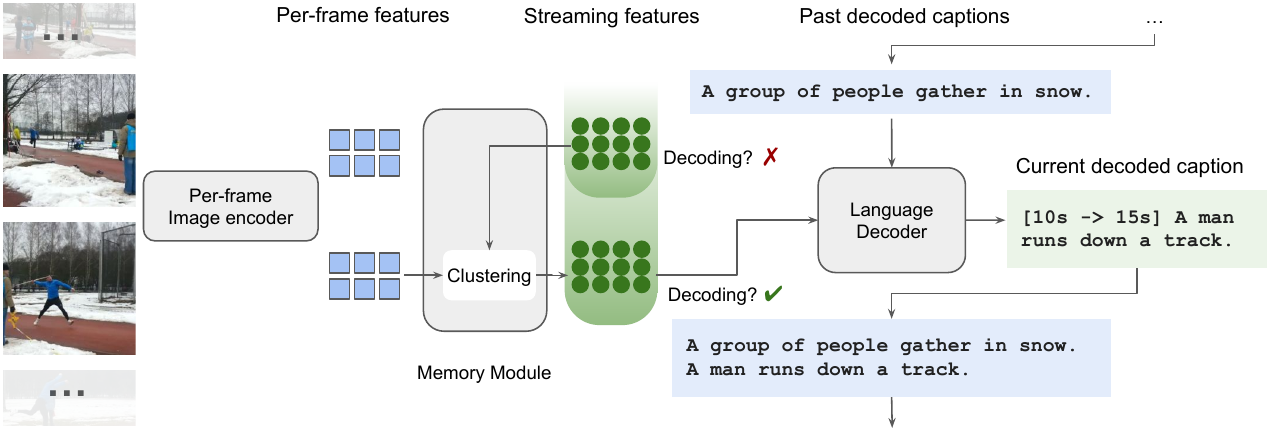}
	\vspace{-3mm}
	\caption{
		\textbf{Illustration of our framework.} Each frame is passed through an image encoder, one at a time.
	A memory model, based on clustering, maintains compressed visual features from the beginning up to the current frame.
	At certain frames, denoted as ``decoding points'', we decode the representations from our memory into captions and their timestamps. 
	Earlier text predictions, if available, are also passed as a prefix to the language decoder for the following decoding points.
	Our model can run on videos of arbitrary length, as the memory has a constant size, and can also output predictions before processing the whole video.
	} 
	\lblfig{framework}
	\vspace{-\baselineskip}
\end{figure*}

Prior work on dense video captioning does not handle either long inputs or long outputs.
Given a video of any length, state-of-the-art models either sample very few frames (e.g., 6 frames~\cite{wang2022git}) with large strides~\cite{wang2022git,chen2023pali,li2023blip} (i.e., temporal downsampling),
or keep one feature per-frame for all the frames~\cite{yang2023vid2seq,wang2021end,zhou2018end} (i.e., spatial downsampling).
With long textual output, current models solely rely on auto-regressive decoding~\cite{graves2013generating} to generate multiple sentences at the end.

In this work, we design a \emph{streaming} model for dense video captioning as shown in Fig.~\ref{fig:teaser}.
Our streaming model does not require access to all input frames concurrently in order to process the video thanks to a memory mechanism.
Moreover, our model can produce outputs causally without processing the entire input sequence, thanks to a new streaming decoding algorithm.
Streaming models such as ours are inherently suited to processing long videos -- as they ingest frames one at a time.
Moreover, as the output is streamed, intermediate predictions are produced before processing the full video.
This property means that streaming models can in theory be applied to process live video streams, as required for applications such as video conferencing, security and continuous monitoring among others.

In order to develop our streaming model,
we first propose a novel memory mechanism, that takes frames once at a time.
The memory model is based on K-means clustering, and uses a fixed number of cluster-center tokens to represent the video at each timestamp.
We show that this method is simple and effective, and can process variable numbers of frames, with a fixed computational budget at decoding.

We also develop a streaming decoding algorithm, and train our network such that given a ``decoding point'' (\reffig{framework}) at a particular timestamp, it predicts all event captions that ended before it given the memory features at that timestamp.
Our network is thus trained to make predictions at any timestamp of the video, and not just at the end of the video as in conventional, non-streaming models.
Furthermore, we provide predictions from earlier decoding points as contexts for later ones.
This context avoids predicting duplicated events, and can be used as an ``explicit'' memory in natural language summarising the earlier video.
Our streaming output is also motivated by the fact that as the video length grows, our memory will inevitably lose information over time as its size is bounded.
We avert this issue by making predictions before we have processed the entire video, and still keep early information via language context.

We evaluate our approach on three popular dense video captioning datasets, ActivityNet~\cite{krishna2017dense}, YouCook2~\cite{zhou2018towards} and ViTT~\cite{huang2020multimodal}.
Our results show that our streaming model significantly improves over the state-of-the-art, which inevitably uses fewer frames or fewer features, by up to $\mathbf{11.0}$ CIDEr points.
We show that our method generalizes across both GIT~\cite{wang2022git} and Vid2Seq~\cite{yang2023vid2seq} architectures. %
Finally, our proposed memory can also be applied to paragraph captioning, improving baselines by $1$-$5$ CIDEr points.

\section{Related Work}

\noindent\textbf{Dense video captioning.}
Dense video captioning requires captioning events, and localizing them temporally.
Traditionally, prior work used a two-stage approach, first localizing events in video, and then subsequently captioning them~\cite{krishna2017dense, iashin2020better, iashin2020multi, wang2018bidirectional, wang2020event}.
More recent end-to-end approaches include PDVC~\cite{zhu2022end} which infers event captions and timestamps using a DETR-like~\cite{carion2020end} model.
Vid2Seq~\cite{yang2023vid2seq} augments the vocabulary of a language model with timestamp tokens, allowing them to generate concatenated event captions in the same manner as a regular captioning model.
We also use the output formulation of~\cite{yang2023vid2seq}, as it integrates well with foundation vision-language models~\cite{wang2022git}.
A related, but different problem, is that of audio description in movies~\cite{soldan2022mad,han2023autoad,han2023autoad2}, which requires generating captions for the visually impaired that must be complementary to speech, and often uses auxiliary, non-causal models to recognize characters or their speech~\cite{han2023autoad2}.

As far as we are aware, all prior dense captioning models are not causal, as they encode the entire video at once.
Moreover, to process long videos, they typically use visual features that are downsampled heavily (by selecting a few frames ~\cite{wang2022git, chen2023pali,li2023blip}, or spatially pooling features per frame~\cite{yang2023vid2seq, wang2021end,zhou2018end}).
In contrast, we process videos in a \emph{streaming} manner, processing long input sequences one frame at a time with a memory module, and streaming output sentences with a novel decoding algorithm.

\noindent{\textbf{Models for long videos.}}
A common way of processing longer videos is to use a memory mechanism to provide a compact representation of past events.
With transformers, memory can easily implemented by using tokens from past observations as inputs for the present time step~\cite{dai2019transformer,wu2022memorizing,petroni2019language}.
Examples in vision include%
~\cite{wu2019long, wu2021towards, pan2021actor, herzig2022object} which pre-extract features offline and retrieve them during inference, and are thus not causal. 
MemViT~\cite{wu2022memvit} uses token activations from previous time steps as inputs to the current time step.
However, this means that the sequence length grows over time and so it cannot handle arbitrarily long videos.

An alternate view of memory is to see it as a method of compressing previously observed tokens into a smaller, fixed-size set to be used at future time steps.
Token Turing Machines~\cite{ryoo2023token} summarize past and current observations using the token summarization module of~\cite{ryoo2021tokenlearner}.
MovieChat~\cite{song2023moviechat} follows a similar idea, but uses a variant of Token Merging~\cite{bolya2022token} instead to perform the summarization.
TeSTra~\cite{zhao2022real} uses an exponential moving average to integrate video features instead.
The advantage of such approaches is that the memory bank has a fixed size, and therefore the computational cost remains bounded regardless of the length of the video.
Our memory model has this same desirable property. However, our memory is based on clustering, using the centers from a K-means-like algorithm to summarize tokens from each time step, and we show experimentally that this outperforms other alternatives.

\noindent\textbf{Causal models in video.}
Our streaming model is causal, meaning that its output only depends on current and past frames, without access to future frames.
Although we are not aware of prior causal models for dense video captioning, there are causal models in many other vision domains.
Online action detection~\cite{de2016online, kondratyuk2021movinets, zhao2022real, zhao2023streaming} aims to predict action labels for videos in real-time without access to future frames.
Analogously, online temporal action localization~\cite{kang2021cag,buch2017sst,singh2017online} models also predict start- and end-times after an action is observed.
Most models for object tracking~\cite{bergmann2019tracking,wang2020towards} and video object/instance segmentation~\cite{li2020towards,milan2016mot16,caelles20192019} are causal too.

A common theme in the above tasks is that the model must make a prediction at each frame of the video.
In contrast, we focus on dense video captioning~\cite{krishna2017dense}, which is challenging as the output captions do not have a one-to-one correspondence with the video frames.
We address this problem by proposing a streaming decoding algorithm.

\section{Streaming Dense Video Captioning}

Given a video $\mathbf{V} \in \mathbb{R} ^ {T \times H \times W \times 3}$, our goal is to produce a set of temporally localized captions: $\{(s, e, \mathbf{c})_1, \ldots, (s, e, \mathbf{c})_{n_e}\}$, where $s \in \mathbb{R}$ and $e \in \mathbb{R}$ are the starting and ending timestamps ($0 \le s < e \le T$), respectively, $\mathbf{c} = [w_1, \cdots, w_n]$ is a sequence of word tokens, and $n_e$ the number of events.
Each word token $w_i$ is an integer in the range $[0, |V|]$, indexing the vocabulary $V$.
We begin by describing conventional captioning models~(Sec.~\ref{sec:method_prelim}), before detailing how we develop a streaming model (Fig.~\ref{fig:framework}) by streaming inputs with memory (Sec.~\ref{sec:method_memory}) and outputs with decoding points (Sec.~\ref{sec:method_streaming-outputs}).

\subsection{Preliminaries}
\label{sec:method_prelim}

Captioning models broadly consist of a vision encoder followed by a text decoder.
We outline these approaches, and show how they can be extended to dense captioning next.

\noindent\textbf{Visual encoder.} The first step is to encode the video into features $\mathbf{f} = \mathcal{F}(\mathbf{V}), \mathbf{f} \in \mathbb{R}^{N \times D}$, where $N$ is the feature resolution (\ie, number of tokens for transformer-based encoders), and $D$ is the feature dimension.
The visual feature encoder $\mathcal{F}$ can be a native video backbone~\cite{arnab2021vivit,bertasius2021space}, or an image encoder~\cite{radford2021learning,yu2022coca} applied to each individual frame.
In the latter case, the video feature is a stack of image features, \ie $N = T \cdot N_f$, where $N_f$ is the number of tokens per frame.
We use a per-frame encoding, but instead of pre-extracting them from the whole video, we use a memory mechanism to process the features in a causal, streaming fashion (Sec.~\ref{sec:method_memory}) that can generalize to longer video durations.

\noindent\textbf{Text decoder.}
Given the visual features, $\mathbf{f}$, and optional textual prefix tokens, $\mathbf{p}$, the text decoder, $\mathcal{D}$ generates a sequence of word tokens, $\mathbf{c}$ from them.
We use an autoregressive~\cite{graves2013generating, vaswani2017attention} decoder that generates the next word token, $w_i$, conditioned on previous words, $\mathbf{w}_{1:i-1}$, and prefix if provided as $w_i = \mathcal{D}(\mathbf{f}, \mathbf{p}, \mathbf{w}_{1:i-1})$.
Note that prefix tokens are typically not used in captioning tasks, but are used in question-answering (QA) to encode the input question.
Concretely, the text decoder, $\mathcal{D}$, is a sequence of transformer layers~\cite{vaswani2017attention} operating on a concatenation of visual features $\mathbf{f}$ and word embeddings of the prefix~\cite{wang2022git,raffel_jmlr_2020}.
This architecture is shown to be effective in both captioning and QA tasks across image and video~\cite{wang2022git, li2023blip, chen2023pali}.

\noindent\textbf{Dense video captioning with timestamps.}
Combining the above visual encoder and text decoder gives a basic architecture for video captioning.
To extend it for captioning multiple events with starting and ending timestamps, Vid2Seq~\cite{yang2023vid2seq} introduced two main modifications:
First, it augments the vocabulary, $V'$, of the captioning model with time tokens, $w^s$ and $w^e$, which represent the starting and ending times, respectively.
A single event is therefore represented as $\mathbf{c}' = [w^{s}, w^{e}, w_1, \cdots, w_n]$, and $|V'| = |V| + |T|$ where $|V| \le w^s < w^e \le |V'|$, and $|T|$ is the number of time tokens. %
Second, Vid2Seq concatenates all timed captions into a single long caption that is ordered by starting time: $\mathbf{C} = [\mathbf{c}'_1, \mathbf{c}'_2, \cdots, \mathbf{c}'_{n_e}]$ where $n_e$ is the number of events.
Therefore, dense video captioning can be formulated as standard video captioning with target $\mathbf{C}$.

Despite its effectiveness, the (generalized) Vid2Seq~\cite{yang2023vid2seq} architecture has a number of key limitations:
First, it forwards visual features from the whole video, $\mathbf{f}$ through the decoder, meaning that it does not scale effectively to longer videos and more tokens.
In addition, as Vid2Seq predicts all event captions at once, after processing the whole video, it struggles with predicting long, detailed captions.
To address these issues, we introduce \textbf{streaming} dense video captioning models, where we process the inputs once at a time using a memory module to bound computational costs, and stream the outputs such that we can make predictions before processing the whole video.

\subsection{Streaming inputs using memory}
\lblsec{method_memory}

The input visual features, $\mathbf{f}$, have dimensionality $\mathbb{R}^{T \cdot N_f \times D}$, where typical values are $T > 64$ for a sparsely sampled (\eg 1 FPS), few-minute-long video, and $N_f\!=\!257$ tokens per-frame for a vision transformer such as CLIP~\cite{radford2021learning}.
Directly feeding all $T\!\cdot\!N_f$ to the text decoder is prohibitively expensive, due to the quadratic complexity of self-attention. %
Therefore, existing methods aggressively downsample $\mathbf{f}$ to reduce the number of tokens (by temporally sampling a few frames with large strides~\cite{wang2022git, chen2023pali}, or spatially subsampling each frame to a single token~\cite{yang2023vid2seq, zhou2018end, wang2021end}).
Even so, we will reach memory limits with longer videos, and the information required for fine-grained localization and description is lost.
Therefore, rather than aggressive downsampling, we use a memory mechanism to process all tokens frame-by-frame, which ensures that the computational cost is bounded irrespective of the length of the video.

\newlength{\textfloatsepsave}
\setlength{\textfloatsepsave}{\textfloatsep}
\setlength{\textfloatsep}{0pt}
\SetNlSty{textnormal}{\scriptsize}{}

\begin{algorithm}[!t]  %
\SetAlgoNlRelativeSize{1}
\footnotesize
\caption{\small{Updating memory tokens at a timestamp.}}
\lblalg{clustering}
\lblalg{tracking}
\SetAlgoLined
\SetKwInOut{Input}{Input}
\SetKwInOut{Output}{Output}
\SetKwInOut{Hyper}{Hyperparameters}
\Input{
    $\mathbf{M}_{t - 1} \in \mathbb{R}^{K \times D}$: memory tokens \\
    $\mathbf{W}_{t - 1} \in \mathbb{R}^{K}$         : weights of memory tokens \\
    $\mathbf{f}_{t} \in \mathbb{R}^{{N_f} \times D}$: incoming tokens \\
}
\Hyper{$\tau$: number of K-means iterations.}
\Output{
    $\mathbf{M}_{t} \in \mathbb{R}^{K \times D}$: updated memory tokens \\
    $\mathbf{W}_{t} \in \mathbb{R}^{K}$: updated weights of memory \\
}
$\mathbf{X} \leftarrow [\mathbf{M}_{t-1}, \mathbf{f}_t]$ \hfill // Concatenate memory and incoming tokens. \\
$\mathbf{W} \leftarrow [\mathbf{W}_{t-1}, \mathbf{1}]$ \hfill // Initialize incoming weights and concatenate. \\
$\mathbf{M}_t \leftarrow \mathbf{M}_{t - 1}$ \hfill // Initialize new cluster centers as the old centers. \\

\For{i $\leftarrow$ 1: $\tau$}{
    $\mathbf{d}$ $\leftarrow$ pairwise\_l2\_distance($\mathbf{X}$, $\mathbf{M_t}$) \hfill // Shape ($K\!\!+\!\!N_f, K$). \\
    $\bm{\delta} \leftarrow \mathbf{d}$.argmin(axis=1) \hfill // Assign each token to a center. \label{alg:kmeans_assignment} \\
    $\bm{\delta} \leftarrow$ make\_onehot($\bm{\delta}$, K) \hfill // Binary. Shape($K\!\!+\!\!N_f, K$). \\
    $\mathbf{W}_t \leftarrow  \bm{\delta}^\top \mathbf{W}_{t}$ \hfill // Compute \#tokens assigned to each center. \\
    $\mathbf{A} \leftarrow \bm{\delta}^\top$ / $\mathbf{W}_t$ \hfill // Weight matrix. ``/'' is elementwise div.\\
    $\mathbf{M}_t \leftarrow \mathbf{A} \mathbf{X}$ \hfill // Compute new centers as a linear function. \\
}

\Return{$\mathbf{M}_{t}$, $\mathbf{W}_{t}$}
\end{algorithm}

Let $K$ be a pre-defined memory size, the memory at each time $\mathbf{M}_t$ would always be the constant size $K$, i.e., $\mathbf{M}_t \in \mathbb{R}^{K \times D}, \forall t$.
We interpret $\mathbf{M}$ as being a summary of all relevant information in the video, and initialize it by taking the first $K$ tokens from $\mathbf{f}$.
Therefore, we set $K$ as a multiple of $N_f$, such that the initial memory is the features of the first $\frac{K}{N_f}$ frames: $\mathbf{M}_{K / N_f} = [\mathbf{f}_1, \cdots, \mathbf{f}_{K / N_f}]$.

Next, we update the memory at each timestamp for each incoming frame $\mathbf{f}_t$.
Our intuition is to keep as much \emph{diverse} information in the original video as possible, while not increasing the storage budget (\ie by keeping a constant memory size $K$).
We thus propose a K-means-like clustering algorithm, to use the feature cluster centers as the approximate video features.
To avoid the cluster centers biasing quickly to incoming features, we keep track of the number of merged tokens in each cluster center.
We use this as a momentum weight, so that cluster centers that are merged from more tokens change slower.
The detailed algorithm diagram is provided in Alg.~\ref{alg:clustering}, and illustrated in \reffig{memory}. %

The K-means algorithm is not differentiable with respect to the assignment of data points to cluster centers,
\textbf{$\bm{\delta}$} (Line \ref{alg:kmeans_assignment} of Alg.\ref{alg:clustering}).
However, the inputs and outputs of our memory module are the updated cluster centers, $\mathbf{M}_t$, which is a linear mapping of the input $\mathbf{X} = [\mathbf{M}_{t - 1}, \mathbf{f}_t]$, as $\mathbf{M}_t = \textbf{A} \mathbf{X}$, where $\textbf{A}$ is a weight matrix computed from $\textbf{X}$.
Therefore, even though we cannot compute the gradient of $\mathbf{A}$ with respect to $\mathbf{X}$, we can compute the gradient of $\mathbf{M}_t$ with respect to the input $\mathbf{X}$, and thus to the input visual feature $\mathbf{f}$.
As a result, we can use our memory module in any part of a neural network, and learn parameters in preceding layers.

\begin{figure}[t]
	\centering
    \includegraphics[width=0.85\linewidth]{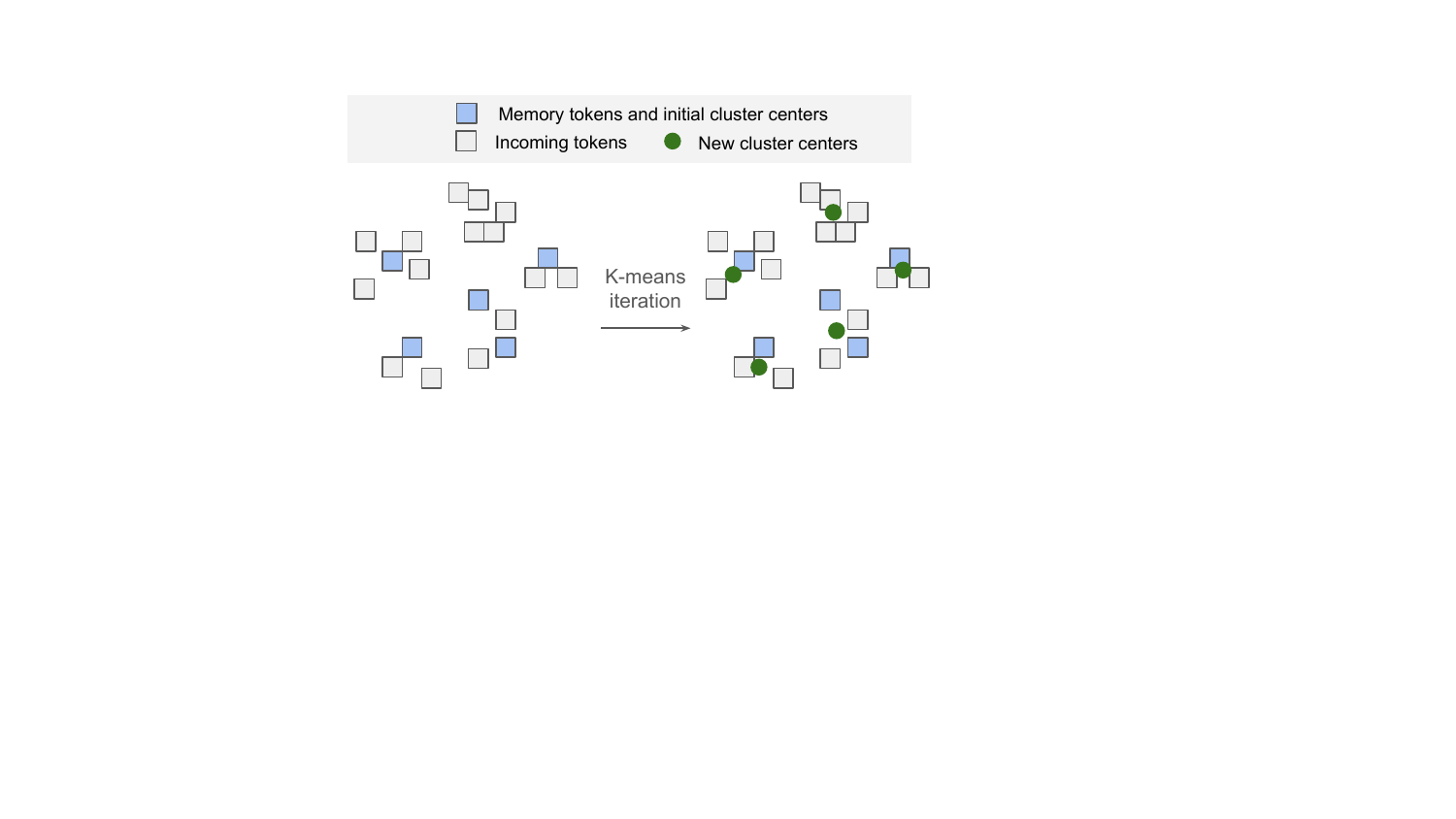}
    \vspace{-1\baselineskip}
	\caption{\textbf{Illustration of our clustering-based memory module.}
		The current memory tokens are shown by \textcolor{bluekmeans}{blue} squares.
		At each time step, the memory tokens evolve by integrating information from the incoming tokens (\textcolor{gray}{gray} squares), using K-means iterations to produce the updated memory tokens (\textcolor{greenkmeans}{green} circles).
	}
	\vspace{\baselineskip}
	\lblfig{memory}
\end{figure}

\subsection{Streaming outputs with decoding points}
\lblsec{method_streaming-outputs}

The memory module from \refsec{method_memory} enables us to efficiently ingest long input videos.
However, it is still desirable for our model's text decoder to predict outputs before it has processed the entire input sequence:
Streaming the output substantially decreases the latency of the model, as we do not have to wait for the model to process the entire input sequence to make predictions.
This is particularly relevant for processing, for example, live video streams.
Furthermore, streaming the output can in fact increase our model's accuracy:
As we have a memory with a fixed size, $K$, from which we decode outputs, we will inevitably lose information over time.
We can therefore avert this issue by making predictions before we have processed the entire video.

\begin{figure}[t]
	\center
    \includegraphics[width=0.85\linewidth]{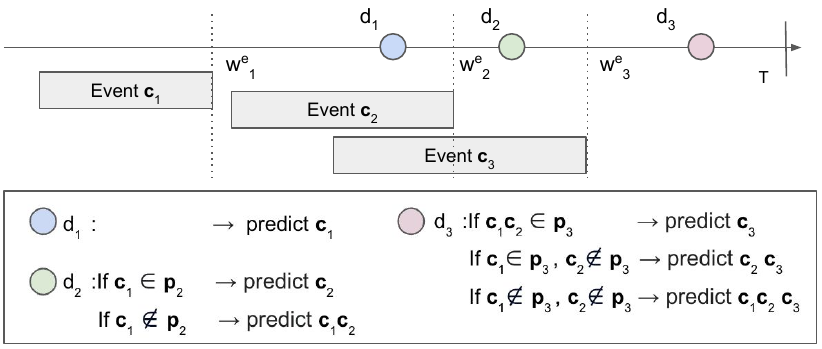}
	\vspace{-3mm}
	\caption{\textbf{Decoding point supervision in training.}
		A decoding point, $d_i$, can be at any frame.
		At each point, we take the memory features, $\mathbf{M}_{d_i}$, and predict all events that have finished before $d_i$, and are not in the prefix $\mathbf{p}$.
        Therefore, the union between the prefix and the prediction target covers all events finished before it.
	}
	\lblfig{method_decoding_points}
	\vspace{\baselineskip}
\end{figure}

\paragraph{Decoding points}
As shown in Fig.~\ref{fig:method_decoding_points}, we define ``decoding points'', $d_i$, as intermediate timestamps where we decode event captions given the features in our memory, $\mathbf{M}_{d_i}$.
We train our model such that at each decoding point, $d_i$, the model predicts all event captions that finished before it.
More specifically,
\begin{equation}
	\mathcal{Y}_i = \{ (w^s_j, w^e_j, \mathbf{c}_j) | w^e_j \leq d_i \},
\end{equation}
where $\mathcal{Y}_i$ is the set of all event captions corresponding to the $i^{\text{th}}$ decoding point $d_i$,
and $w^s_j$, $w^e_j$ are the starting and ending time of the $j^{\text{th}}$ event.

As decoding points are applied sequentially, later decoding points should have access to the predictions of earlier decoding points, and should not repeat them again.
Therefore, from the second decoding point onwards, we concatenate the outputs of previous decoding points as the prefix to the text decoder, as shown in Fig.~\ref{fig:method_decoding_points}.
Moreover, during training, we perform further data augmentation by randomly removing some of the previous event captions from the prefix, and adding them to the target instead, to increase robustness to potential errors in earlier predictions.
We therefore denote our prefixes and captioning targets during training as
\begin{align}
	\mathbf{p}_i &= [\mathbf{c}'_1, \mathbf{c}'_2, \ldots, \mathbf{c}'_{j-1}] \label{eq:streaming_prefix} \\
	\mathbf{y}_i &= [\mathbf{c}'_j, \mathbf{c'_{j + 1}}, \ldots, \mathbf{c}'_{|\mathcal{Y}_i|}],
\end{align}
where $j < |\mathcal{Y}_i|$ is a randomly chosen splitting point to partition the target and context.
During inference, we use the models actual predictions, $\mathbf{p}_i = [\hat{\mathbf{y}}_1, \hat{\mathbf{y}}_2, \ldots, \hat{\mathbf{y}}_{i-1}]$, instead.

In practice, we uniformly sample decoding points during both training and inference, with a stride of $S$ frames.
Since the model is trained to predict all event captions before the decoding point, it means that the exact location at inference time does not need to match the event boundaries closely.
The number of decoding points can be also different between training and inference, as we will ablate.
This method is both simple and scalable, as we show experimentally in the next section.

\section{Experimental Evaluation}

\subsection{Experimental Setup}

\subsubsection{Datasets and Evaluation Metrics}
We evaluate our model on the three most popular dense video captioning datasets:
ActivityNet Captions~\cite{krishna2017dense}, YouCook2\cite{zhou2018towards}, and ViTT~\cite{huang2020multimodal}.

\noindent\textbf{ActivityNet Captions}~\cite{krishna2017dense} contains 8,649 training videos and 4,267 validation videos (considering videos that are still online on YouTube).
The videos are untrimmed, each containing multiple action events, with an average video length of 2 minutes and an average of 3.7 events.
The videos are selected from the ActivityNet~\cite{caba2015activitynet}, which covers a wide range of visual domains.
Each video is carefully labeled by human annotators.
We use this dataset for both dense captioning (predicting a sequence of captions with their start- and end-times) and paragraph captioning (predicting the aforementioned captions as a single paragraph without timestamps).

\noindent\textbf{YouCook2}~\cite{zhou2018towards} contains 1,333 training videos and 456 validation videos.
The videos are on average $5.3$ minutes long, with an average of $7.8$ events.
This dataset focuses on cooking scenes, and most videos contain one main actor cooking while describing the recipe.
The videos are manually annotated, aided by speech transcriptions obtained through Automatic Speech Recognition (ASR).
As expected due to the annotation process, using ASR as an additional modality can be used to improve model predictions~\cite{yang2023vid2seq}.
However, our work focuses on using visual-only inputs, to study our proposed streaming model.
We also perform dense and paragraph captioning on this dataset.

\noindent\textbf{ViTT}~\cite{huang2020multimodal} contains 4,608 training videos and 2,301 testing videos, which are on average $4.7$ minutes long and contain $7$ events per video.
It is collected in a similar way as YouCook2, but focuses on more diverse domains.
Again, we do not use the ASR inputs in our experiments.

\noindent\textbf{Evaluation Metrics}
For all datasets, we report standard dense captioning evaluation metrics following Vid2Seq~\cite{yang2023vid2seq} and PDVC~\cite{wang2021end}.
The primary metric is CIDEr which has been adapted to jointly evaluate captioning and localisation.
In particular, we average the CIDEr score~\cite{vedantam2015cider} for positive ground-truth-prediction pairs with a temporal IoU above the thresholds, $\{0.3, 0.5, 0.7, 0.9\}$.
Similarly, we report METEOR~\cite{banerjee2005meteor} averaged over multiple IoU thresholds.
Finally, we report the recently proposed SODA$_c$~\cite{fujita2020soda}, which considers all event captions from the same video, and is therefore a more holistic measure.

\subsubsection{Implementation details}
We implement our streaming model on two video captioning architectures, GIT~\cite{wang2022git} and Vid2Seq~\cite{yang2023vid2seq}.
Both use a ViT-L~\cite{dosovitskiy_iclr_2021} initialized from CLIP~\cite{radford2021learning} as image encoder $\mathcal{F}$.

GIT~\cite{wang2022git} concatenates all tokens from all input frames, $\mathbf{f}$, and feeds them to a 6-layer transformer language decoder.
We apply our streaming input module before the language decoder, \ie, rather than concatenating frame features, we use our memory features instead.
The original GIT paper pretrains the language decoder on in-house data~\cite{wang2022git}.
As we do not have access to the original data or weights, we pretrain our GIT model on the WebLI dataset~\cite{chen2022pali}. %

Vid2Seq~\cite{yang2023vid2seq} pools all frame tokens spatially into 1 token per frame, and then applies a 12-layer temporal transformer~\cite{arnab2021vivit} to obtain the final visual features.
The language decoder is a 12-layer T5-Base decoder~\cite{raffel_jmlr_2020}.
Vid2Seq pretrains the temporal transformer and the T5 decoder on narrated videos from the YT-Temporal dataset~\cite{zellers2022merlot}. We use their public, pretrained weights as our initialization.
We apply our streaming input module before the temporal transformer, to ensure the intermediate features remain causal.

When finetuning our model for each dataset, we freeze the frame encoder following prior works~\cite{yang2023vid2seq,wang2021end,zhu2022end}.
We provide detailed training hyperparameters in the supplementary.
For each architecture, the training hyperparameters are shared across the 3 datasets.

\subsection{Analysis of Streaming Modules}
\lblsec{exp_streaming_model_analysis}

We first analyze each of the main components of our model: streaming the input with our memory module~(Sec.~\ref{sec:method_memory}), and then streaming outputs using decoding points~(Sec.~\ref{sec:method_streaming-outputs}).
Unless otherwise stated, we use the GIT~\cite{wang2022git} backbone for our experiments.
Due to randomness in training and evaluation, for all ablation experiments (\reftbl{memory}-\reftbl{streaming_output_point_ablation}), we repeat the same run 3 times and report the averaged metrics.
When compared to other methods (\reftbl{streaming_model}, \reftbl{paragraph}), we report results for the best model.

\subsubsection{Streaming inputs with memory}

\begin{table}[!t]
\centering
\vspace{-0.5\baselineskip}
\begin{tabular}{@{}l@{}c@{\ \ }c@{\ \ }c@{\ \ }c@{\ \ }c@{}}
\toprule
      & \#Tokens & $T\!\!=\!\!16$                & $T\!\!=\!\!32$                & $T\!\!=\!\!64$                & $T\!\!=\!\!128$               \\
\midrule
No memory         & $T\!\!\times\!\!N_f$ & \textbf{29.8} & -                        & -                        & -                        \\
Spatial pooling   & $T$ &{27.6} & {27.3} & {27.9}   & {27.4} \\
Temporal pooling  & $N_f$ &{29.3} & {28.0} & {26.8}   & {25.2} \\
EMA~\cite{zhao2022real}        & $N_f$       &               28.2           &               26.3           &          22.0                &          16.3                \\
MovieChat~\cite{song2023moviechat}         & $K$ & {29.3} & \textbf{29.2} & {28.9} & {29.3} \\ \midrule
Clustering (ours) & $K$   &       \textbf{29.8}                   & \textbf{29.2} &       \textbf{30.6} & \textbf{30.4}                 \\
\bottomrule
\end{tabular}
\vspace{-3mm}
\caption{\textbf{Ablation on memory modules.}
We show CIDEr (averaged over multiple IoUs) on ActivityNet with GIT under different \# input frames $T$.
The 2nd column shows the number of input tokens to the language decoder.
$N_f\!\!=\!\!257$ is the number of tokens per-frame. $K\!\!=\!\!514$ is the number of memory tokens.
Our module benefits from more frames, and outperforms other alternatives.
}
\vspace{\baselineskip}
\label{tab:memory}
\end{table}

\begin{table}[t]
    \begin{subfigure}[t]{0.32\linewidth}
    \vskip 0pt
    \centering
        \begin{tabular}{@{}l@{\ \ \ \ }c@{\ }}
        \toprule
        $K$            & CIDEr \\
        \midrule
        $257$                  & 29.4                         \\
        $257 \times 2$       & \textbf{30.6}                      \\
        $257 \times 3$ & 29.8                      \\
        \bottomrule
        \end{tabular}
        \caption{\#memory tokens.}
    \end{subfigure}
    \hfill
    \begin{subfigure}[t]{0.32\linewidth}
    \vskip 0pt
    \centering
        \begin{tabular}{@{}l@{\ \ \ \ }c@{\ }}
        \toprule
        Iters. & CIDEr  \\
        \midrule
        1                  &  30.3 \\
        2                  &  \textbf{30.6}     \\
        4                  &  30.3 \\
        \bottomrule
        \end{tabular}
        \caption{K-means iterations.}
    \end{subfigure}
    \hfill
    \begin{subfigure}[t]{0.32\linewidth}
    \vskip 0pt
    \centering
        \begin{tabular}{@{}l@{\ }c@{\ }}
        \toprule
        Momentum   & CIDEr  \\
        \midrule
        \xmark                  &  29.7 \\
        \cmark                  &  \textbf{30.6}     \\
        \bottomrule
        \end{tabular}
        \caption{Momentum term in clustering}
    \end{subfigure}
    \vspace{-3mm}
    \caption{\textbf{Ablation on memory module hyperparameters.} We perform experiments on ActivityNet with GIT with $64$ frames. %
    }
    \label{tab:memory_parameters}
    \vspace{\baselineskip}
\end{table}

As the GIT~\cite{wang2022git}
architecture concatenates visual features, $\mathbf{f}$, from multiple frames before passing it to the language decoder, it is limited by the number of frames that it can process.
Our clustering-based memory module allows us to process more frames, and we consider the following approaches to evaluate it:

\begin{table}[!t]
        \vspace{-0.5\baselineskip}
	\centering
	\begin{subfigure}[t]{0.95\linewidth}
		\vskip 0pt  %
		\centering
		\scalebox{0.8}{
			\begin{tabular}{l@{\ \ }c@{\ \ \ \ }c@{\ \ \ \ }c@{\ \ \ \ }c@{\ \ \ \ }c@{\ \ \ \ }c}
				\toprule
				Number of decoding points & 1 & 2 & 4 & 8 & 16 & 20\\
				\midrule
				CIDEr & 30.6 & 34.5 & 38.8 & 39.5 & \textbf{40.6} & 40.4 \\
				\bottomrule
			\end{tabular}
		}
		\caption{Number of decoding points during training.}
		\label{tab:ablation_decoding_points_num}
	\end{subfigure}
	\begin{subfigure}[b]{0.32\linewidth}
		\vskip 0pt
		\centering
            \setlength{\tabcolsep}{2pt}
		\scalebox{0.8}{
			\begin{tabular}{@{}lc@{}}
				\toprule
				Prefix               & CIDEr   \\
				\midrule
				None                   &    23.1 \\
				Captions             &    \textbf{40.6}  \\
				Captions \& time &   39.3    \\
				\bottomrule
			\end{tabular}
		}
		\caption{%
			Context provided after the first decoding point.
		}
		\label{tab:ablation_decoding_points_prefix}
	\end{subfigure}
	\hfill
	\begin{subfigure}[b]{0.32\linewidth}
		\vskip 0pt
		\centering
            \setlength{\tabcolsep}{2pt}
		\scalebox{0.8}{
			\begin{tabular}{@{}l@{\ \ \ \ \ \ }c@{}}
				\toprule
				Aug. Prefix               & CIDEr \\
				\midrule
				\xmark & 37.6 \\
                    \checkmark & 40.6 \\
                \\
				\bottomrule
			\end{tabular}
		}
		\caption{
            Random masking prefix as augmentation during training.
		}
		\label{tab:aug_prefix}
	\end{subfigure}
	\begin{subfigure}[b]{0.32\linewidth}
		\vskip 0pt
		\centering
            \setlength{\tabcolsep}{2pt}
		\scalebox{0.8}{
			\begin{tabular}{@{}l@{\ \ \ \ \ \ }c@{}}
				\toprule
				Stride               & CIDEr \\
				\midrule
				$32$ & 40.6 \\
                    $21$ & 34.7 \\
                \\
				\bottomrule
			\end{tabular}
		}
		\caption{
			Decoding point stride during inference (input 64 frames).
		}
		\label{tab:ablation_decoding_points_location}
	\end{subfigure}
    \vspace{-3mm}
	\caption{\textbf{Ablation on streaming outputs.}
		(a) Increasing the number of decoding points during training consistently improves accuracy, as it provides more supervision to the model.
		(b) It is critical to provide context after the first decoding point, so that the model does not repeat predictions. Providing captions alone, without timestamps is sufficient.
            (c) It is important to provide imperfect prefixes during training, to mimic model behavior in inference.
            (d) Stride used for decoding during inference. 
		\label{tab:streaming_output_point_ablation}
	}
    \vspace{\baselineskip}
\end{table}

\noindent\textbf{No memory}:
We simply concatenate all visual tokens from all frames as done in GIT~\cite{wang2022git}.
Due to memory constraints, we can only feed up to 16 frames to the decoder.

\noindent\textbf{Spatial- or temporal-pooling}: We pool the visual features, $\textbf{f}$, along either the spatial or temporal dimensions to reduce the number of tokens fed to the language decoder.

\noindent\textbf{EMA}: We use an exponential moving average of frame features, $\mathbf{f}_t$, at each time step, following TeSTra~\cite{zhao2022real}.
We sweep the decay rate from $\{0.9, 0.99, 0.999\}$, finding $0.9$ to perform best.

\noindent\textbf{MovieChat}~\cite{song2023moviechat}.
Finally, the recent MovieChat paper maintains a memory of $K$ tokens.
For each incoming frame, it sequentially processes each token, and merges the two most similar tokens in the memory bank such that the size remains fixed at $K$.
We implement this method using the author's public code.

\begin{table*}[!t]
	\centering
	\begin{tabular}{@{}l@{\ }c@{\ }c@{\ }c@{\ }c c@{\ }c@{\ }c@{\ }c c@{\ }c@{\ }c@{\ }c@{}}
		\toprule
		& \multicolumn{4}{c}{ActivityNet} & \multicolumn{4}{c}{YouCook2} & \multicolumn{4}{c}{ViTT}     \\
		& CIDEr & SODA    & Meteor  & F1       & CIDEr & SODA & Meteor  & F1         & CIDEr & SODA  & Meteor  & F1   \\
		\cmidrule(r){1-1}
		\cmidrule(r){2-5}
		\cmidrule(r){6-9}
		\cmidrule(){10-13}
		MT~\cite{zhou2018end} & 9.3 & --  & 5.0 &   --    &  6.1 & -- & 3.2 &  --      & -- & -- & -- & -- \\
		E2ESG~\cite{zhu2022end} & -- & -- & -- &    --    & 25.0 & -- & 3.5 &  --     & -- & -- & -- & -- \\
		PDVC~\cite{wang2021end} & 29.3 & {6.0} & 7.6 & -- & 28.9 & 4.9 & 5.7 & --     & -- & -- & -- & -- \\
		GIT~\cite{wang2022git}                            & 29.8 & 5.7  & 7.8  & 50.6 & 12.1  & 3.1  & 3.4  & 17.7  & 15.1  & 7.1  & 3.4 & 32.5 \\
		Vid2Seq$^{\dagger}$~\cite{yang2023vid2seq}        & 30.2 & 5.9  & 8.5  & 51.8 & 25.3  & 5.7  & 6.4  & 23.5  & 23.0  & 9.8  & 5.0 & {\bf 37.7} \\
		\cmidrule(r){1-1}
		\cmidrule(r){2-5}
		\cmidrule(r){6-9}
		\cmidrule(){10-13}
		Streaming GIT (ours)     & \textbf{41.2}   & \textbf{6.6}   & 9.0 & 50.9       & 15.4  & 3.2  & 3.6 & 16.6       & 18.5  & 8.3  & 4.0 & 33.9 \\
		Streaming Vid2Seq (ours) & 37.8   & 6.2   & \textbf{10.0}  & {\bf 52.9}      & \textbf{32.9}  & \textbf{6.0}  & \textbf{7.1}  & \textbf{24.1}  & \textbf{25.2}  & \textbf{10.0}  & {\bf 5.8} & 35.4 \\
		\bottomrule
	\end{tabular}
    \vspace{-3mm}
	\caption{\textbf{Comparison to the state-of-the-art on dense video captioning}
		We add our streaming model to both GIT~\cite{wang2022git} and Vid2Seq~\cite{yang2023vid2seq}, denoted by Streaming GIT and Streaming Vid2Seq, respectively, achieving consistent and substantial improvements across three datasets.
		All models use only visual inputs. $^\dagger$denotes version with visual-only inputs.
	}
    \vspace{-3mm}
	\label{tab:streaming_model}
\end{table*}

\reftbl{memory}
compares the results of the different memory modules.
For $T\!\!=\!\!16$, where we can feed all the tokens from the vision backbone, $\mathbf{f}$, into the decoder, ``no memory'' and our method both performs the best.
We expected ``no memory'' to perform well, as it uses the most tokens, $T\!\!\times\!\!N_f$.
However, our clustering-based method achieves the same performance, which suggests that it is able to effectively capture the relevant information in the video with $8 \times$ fewer tokens, and is causal too.
It is not possible to use ``no memory'' for $T\!\!>\!\!16$ due to its computational cost.

With more frames, na\"ively pooling along the spatial-, or temporal-dimensions actually performs worse.
This is likely because we are averaging out information over longer temporal durations, and thus losing the details required for more detailed localization or captioning.
Our method and MovieChat on the other hand, are able to leverage more frames to improve performance, as they keep diverse features within the memory.
Finally, our clustering method outperforms MovieChat and other memory modules for all numbers of frames that we considered, which is why we use it for all future experiments.

We also ablate the hyperparameters of our memory module in Tab.~\ref{tab:memory_parameters}.
Following this experiment, we set $K\!=\!257\!\times\!2$, or the number of tokens in two frames, and use 2 iterations of K-means, as it achieves the best performance.
Our proposed momentum term (Sec.~\ref{sec:method_memory}), which prevents cluster centers from becoming too biased towards incoming frame tokens, also improves CIDEr from 29.7 to 30.6.

\vspace{-3mm}
\subsubsection{Streaming outputs}

We now analyze the effect of our streaming decoding method (Sec.~\ref{sec:method_streaming-outputs}), which enables us to predict event captions from intermediate features in our memory, and utilize previous prediction as the prefix to our text decoder.
Table~\ref{tab:streaming_output_point_ablation} analyses the effects of the number of decoding points during training, the impact of the prefix, and how we select decoding points during inference.

Table~\ref{tab:ablation_decoding_points_num} shows that we achieve significant improvements by increasing the number of decoding points during training, improving the CIDEr score by $10$ points, or 33\% relative, compared to only making predictions at the end, as in conventional models.
Streaming the output with decoding points can provide performance benefits for multiple reasons:
First, as we have multiple decoding points, the required output caption is shorter at each decoding point, making the captioning task easier.
Second, the visual features from our memory, $\mathbf{M}$, may be aligned better with the target text, since $\mathbf{M}$ does not represent the visual features from the entire video as in baseline approaches, but only the features up to the decoding point.
Finally, training with decoding points provides a stronger training signal to the model, as we provide training supervision at each decoding point.
Moreover, it should aid generalization, as the network is being trained to make consistent predictions from more points along the timeline.

\reftbl{ablation_decoding_points_prefix} shows that it is essential to provide past predictions as the prefix.
Without a prefix, the model is poor, underperforming our non-streaming baseline of only decoding at the end ($30.6$).
This is because the model predicts duplicated predictions at each decoding point.
Providing past captions outperforms this baseline substantially.
We find that also adding previously predicted timestamps, does not improve over captions alone, suggesting that temporal boundaries of past events are not particularly informative for future events.
\reftbl{aug_prefix} further shows that while training with a prefix, it is important to mimic inference behavior by including missing captions in earlier predictions.

Table~\ref{tab:ablation_decoding_points_location} examines the choice of decoding points during inference.
We find that using a stride, $S\!=\!32$ (which equates to a decoding point at the middle and end of a 64-frame clip), performs considerably better than $S\!=\!21$ (three uniformly chosen points).
We observed qualitatively that using too many decoding points during inference can sometimes still result in the model making duplicate predictions (even with past prediction as the prefix).
Whilst it is also possible to remove duplicate predictions with non-maximal suppression (NMS~\cite{felzenszwalb2009object}), it is more challenging for captioning models as we also require a calibrated score for each event caption to perform NMS.
We therefore leave an investigation of NMS for dense video captioning to future work.

\begin{figure*}[t]
    \vspace{-0.5\baselineskip}
	\center
    \includegraphics[width=0.98\linewidth]{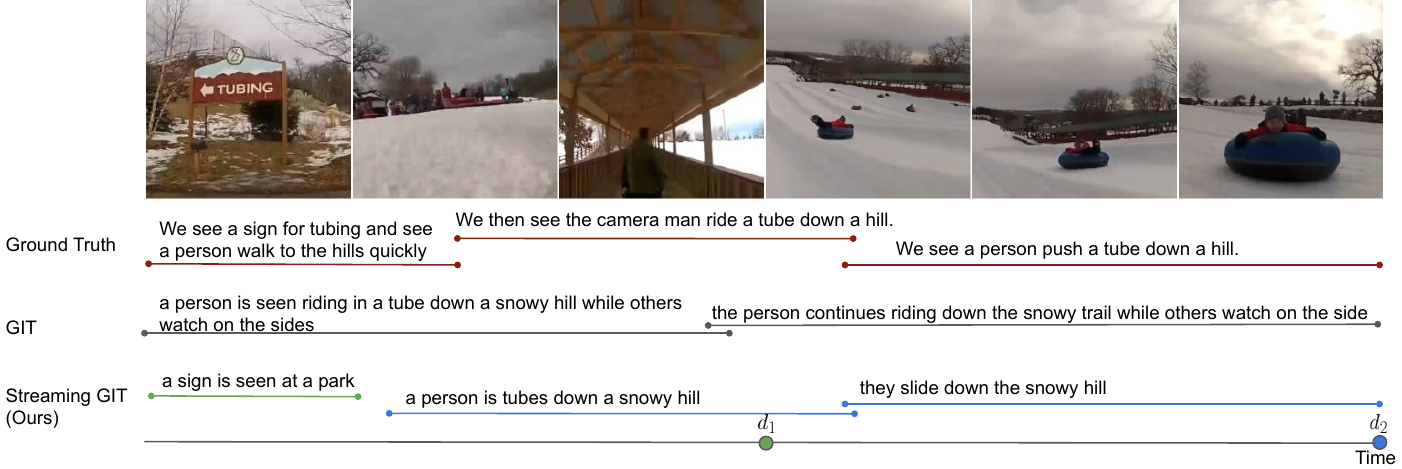}
	\vspace{-4mm}
	\caption{
	\textbf{Qualitative results on ActivityNet validation.}
    Results from the ground truth (top), the baseline (middle), and our model (bottom). We show outputs from two decoding points in green and blue respectively. %
    Our model captures more details than the baseline.
	} 
	\lblfig{qualitative}
    \vspace{-3mm}
\end{figure*}

\vspace{-3mm}
\subsubsection{Generalization to backbones and datasets}
To show the generality of our method, we add our streaming modules onto both the GIT~\cite{wang2022git} and Vid2Seq~\cite{yang2023vid2seq} architectures, which we denote as Streaming GIT and Streaming Vid2Seq, respectively.

{The last two rows} of Tab.~\ref{tab:streaming_model} shows that we improve substantially over both baselines consistently on three datasets.
Our GIT baseline can process a maximum of $N_f\!=\!16$ frames due to memory limitations, and our improvement also stems from the fact that we use $N_f\!=\!64$ for our Streaming GIT thanks to our memory module.
Vid2Seq pools visual tokens spatially such that it only uses a single token per frame.
Therefore, our Streaming Vid2Seq does not use more frames than the baseline, and the improvement is due to our streaming of output event captions, which improves performance significantly as shown in Tab.~\ref{tab:streaming_output_point_ablation}.

Finally, we observe that our Vid2Seq baseline performs substantially better than the GIT baseline on YouCook2.
This difference is due to the pretraining: We used the public, pretrained Vid2Seq checkpoint~\cite{yang2023vid2seq} which was pretrained on YT-Temporal~\cite{zellers2022merlot} -- a dataset with a similar domain to YouCook2.
Note that this is the same experimental protocol as Vid2Seq~\cite{yang2023vid2seq}, the current state-of-the-art.

\subsection{State-of-the-art Comparison}
\lblsec{exp_sota}

Tab.~\ref{tab:streaming_model} also compares our method to the state-of-the-art among dense video captioning methods using only video frames as inputs.
We achieved substantial gains over prior, published works, notably improving CIDEr on ActivityNet by 11.0 points, and YouCook2 by 4.0 points, respectively.
We also achieved improvements, albeit smaller, on SODA and Meteor.
Our improvements on localization (F1) are smaller, showing the gains are more from better captioning qualities.
Fig.~\ref{fig:qualitative} visualizes an example on ActivityNet.

We note that it is possible to further improve results, particularly on YouCook2, by using Automatic Speech Recognition (ASR) as an additional input modality~\cite{yang2023vid2seq}.
This is primarily because the spoken utterances of the actors are well aligned with the visual content, and because ASR was used in the annotation procedure of the dataset itself.
However, integrating multiple modalities such as ASR is orthogonal to the main focus of this work.

\begin{table}[]
\centering
\begin{tabular}{@{}l@{\ \ \ \ \ \ \ \ }c@{\ \ \ \ \ \ \ \ }c@{}}
\toprule
                      & \multicolumn{1}{l}{ActivityNet} & YouCook2           \\
\midrule
MFT~\cite{xiong2018move}                   & 19.1                                  &           -               \\
PDVC~\cite{wang2021end}                  & 20.5                                  &             -            \\
Vid2Seq~\cite{yang2023vid2seq}$^\dagger$  & 28.0                                    &  26.6 \\
GIT~\cite{wang2022git} & 32.5                                  &            28.4             \\
\midrule
Ours
    & \textbf{33.4}                                  &        \textbf{33.9}  \\
\bottomrule
\end{tabular}
\vspace{-3mm}
\caption{\textbf{State-of-the-art comparison on paragraph captioning.}
We report CIDEr on ActivityNet and YouCook2 validation set.
We compare models that do not require action proposal inputs and only takes visual inputs.
$^\dagger$denotes version with visual-only inputs.
Our model achieves the best performance on both datasets.
}
\lbltbl{paragraph}
\end{table}

\noindent\textbf{Paragraph captioning.}
In addition to dense captioning, we also compare with state-of-the-art models on the same datasets for paragraph captioning, which aims to predict the captions throughout the entire video, but without any timestamps.
Therefore, we only apply our streaming input model here, as the timestamps needed to assign decoding points during training are not available in this setting.

We train our Streaming GIT model for paragraph captioning, on both ActivityNet~\cite{krishna2017dense} and Youcook2~\cite{zhou2018towards}.
\reftbl{paragraph} shows that we achieve state-of-the-art results on this task too.
The GIT baseline is our model trained on the same setting as our full model, but uses 16 input frames with all tokens concatenated for the decoder.
This baseline already outperforms the state-of-the-art from visual-only inputs. %
Our model uses more input frames (64 frames), and further boosts the performance by $0.9$ and $5.5$ points on the two datasets, respectively, showing the benefits of our memory module which is consistent with our results in Tab.~\ref{tab:memory}.

\section{Conclusion and Future Work}

We have proposed a streaming model for dense video captioning with two novel components:
A clustering-based memory that can efficiently handle arbitrarily long videos with bounded computation, and a streaming decoding algorithm that enables our model to make predictions before the entire video has been processed.
We achieve this streaming ability while also improving the state-of-the-art on five dense- and paragraph-captioning tasks.

Future work is to develop a benchmark for dense video captioning which requires reasoning over longer videos than current datasets, to better evaluate the abilities of streaming models such as ours.

\small \noindent \textbf{Acknowledgments.} We thank Chen Sun for helpful discussions.

{
    \small
    \bibliographystyle{ieeenat_fullname}
    \bibliography{bibliography}
}

\section*{Appendix}
\appendix

\ifarxivversion
    We provide further training details of our method in Sec.~\ref{sec_supp:training_hyperparams}.
\fi

\section{Training hyperparameters}
\label{sec_supp:training_hyperparams}

All our experiments are conducted using the Scenic library~\cite{dehghani2021scenic} and JAX~\cite{jax2018github}.
With the GIT~\cite{wang2022git} architecture, we first pretrain on the WebLI~\cite{chen2023pali} dataset for general image captioning.
WebLI~\cite{chen2023pali} contains 100M image-text pairs derived from alt-text from the internet.
The image encoder is initialized from CLIP-L~\cite{radford2021learning}, and the language decoder is randomly initialized.
During pretraining, we use the standard label-smoothed (factor $0.1$) cross-entropy loss following GIT~\cite{wang2022git} and train for 10 epochs.
We use the Adam~\cite{kingma2014adam} optimizer, with no weight decay.
The learning rate is set to $5 \times 10^{-5}$ with a batch-size of $1024$, with a cosine decay schedule.
Following GIT~\cite{wang2022git}, we use $0.2\!\times$ lower learning rate for the image encoder.

When finetuning on dense-video captioning datasets~\cite{krishna2017dense,zhou2018towards,huang2020multimodal},
we freeze the image encoder.
We again use the Adam~\cite{kingma2014adam} optimizer with $0$ weight decay.
We train for 20 epochs with batch size of $32$, and use a learning rate of $10^{-5}$, dropped by $10\!\times$ at the 16th epoch.

With Vid2Seq~\cite{yang2023vid2seq}, we take the publicly released pretrained checkpoint\footnote{\url{https://github.com/google-research/scenic/tree/main/scenic/projects/vid2seq}}, which is pretrained on the YT-Temporal dataset~\cite{zellers2022merlot} with a denoising and a captioning objective~\cite{yang2023vid2seq}.
When finetuning on dense-video captioning datasets~\cite{krishna2017dense,zhou2018towards,huang2020multimodal},
we follow their official training parameters.
Specifically, we freeze the image encoder and pool the image tokens among the spatial dimensions to get one token per frame.
The T5~\cite{raffel_jmlr_2020} decoder uses a dropout rate of $0.1$.
We again use Adam~\cite{kingma2014adam} optimizer with $0$ weight decay.
We train for 40 epochs with batch-size 32, and use a learning rate of $3 \times 10^{-4}$ with a cosine decay schedule.

For all models, we follow the standard protocol to use beam-search decoding, with a beam size of $4$ and a brevity penalty of $0.6$~\cite{raffel_jmlr_2020}.
We also emphasize that wherever applicable, all base architectures and backbones are consistent between comparisons and baselines. 

\end{document}